\documentclass[submission,copyright,creativecommons]{eptcs}
\providecommand{\event}{ICLP 2019} 
\usepackage{underscore}           
\usepackage[utf8]{inputenc}
\usepackage{breqn}

\title{Towards Ethical Machines Via Logic Programming}
\author{Abeer Dyoub \quad Stefania Costantini
	\institute{{DISIM, }
	{University of L'Aquila, Italy}}
	\email{abeer.dyoub@graduate.univaq.it \quad stefania.costantini@univaq.it}
	\and
	Francesca A. Lisi
	\institute{DIB and CILA, University of Bari, Italy}
	\email{\quad FrancescaAlessandra.Lisi@uniba.it}
}
\def\titlerunning{Towards Ethical Machines Via Logic Programming}
\def\authorrunning{A.\,Dyoub and S.\,Costantini and F.\,A. Lisi}

\begin{document}
\maketitle

\begin{abstract}
Autonomous intelligent agents are playing increasingly important roles in our lives. They contain information about us and start to perform tasks on our behalves. Chatbots are an example of such agents that need to engage in a complex conversations with humans. Thus, we need to ensure that they behave ethically. In this work we propose a hybrid logic-based approach for ethical chatbots.
\end{abstract}
\section{Introduction}
\label{intro}

Machine ethics is a newly evolving field aiming at creating machines able to compute and choose the best moral action. However the overall aim is not only important for equipping machines with capabilities of moral reasoning, but also for helping us to better understand morality through creating and testing computational models of ethical machines that follow a set of ideal ethical principles.
Since the beginning of this century there were several attempts for implementing ethical decision making into intelligent autonomous agents using different approaches. But, no fully descriptive and widely accepted model of moral judgment and decision-making exists.

In this work we propose a hybrid logic-based approach for modeling ethical machines, particularly ethical chatbots. As a matter of fact the potential of logic programming (LP) to model moral machines was envisioned by Pereira and Saptawijaya \cite{PereiraS16}. In their work the authors investigated the potential of LP for modeling different morality aspects that appear to be amenable to computational modeling by exploiting LP features. 

Chatbot, or virtual assistant, is a computer program or an artificial intelligent software which can simulate a conversation with a user in natural language via auditory or textual methods. They are typically used in dialogue systems for various practical purposes including customer service or information acquisition. From a technological point of view, a chatbot only represents the natural evolution of question answering system leveraging Natural Language Processing.
Businesses are rapidly moving towards the need for chatbots and other self-service technology.
Many banks and insurers, media and e-commerce companies, airlines and hotel chains, retailers, health care providers, government entities and restaurant chains have used chatbots to answer simple questions, increase customer engagement, for promotion, and to offer additional ways to order from them.

However the chatbots raise many ethical questions from privacy to data ownership, to abuse and transparency. Ethics form the foundation of how a chatbot is built, and more importantly, they dictate how a bot interacts with users and how a bot behaves has the potential to influence how an organization can be perceived and unethical behavior can lead to consumer mistrust and litigation issues. Ethical chatbots can promote brand loyalty and help boost profit margins.
The behavior of these machines should be guided by the company's codes of ethics and conduct. However, building such ethical chatbots is not an easy task. Codes of ethics and conduct are mostly abstract rules that lack clear directions for decision making. Customer service codes such as confidentiality, accountability, empathy, fidelity, honesty, etc. are quite difficult to apply in real world situations, they cover a wide range of specific cases. They are subject to interpretations and carry different meanings in different situations. 
It is extremely difficult to use deductive logics alone to build such ethical machines, because it is impossible for experts to define intermediate rules to cover all possible situations. For the sake of transparency and for future ethical decisions, we need detailed ethical principles in place to guide the behavior of our chatbot. To achieve this we need to incorporate a learning technique in the design of our agent.

The approach proposed in this work combine deductive(rule-based) logic programming and inductive(learning) logic programming approaches in one framework for building our ethical agent. We use Answer Set Programming (ASP) for knowledge representation and reasoning, and Inductive Logic Programming (ILP) as a machine learning technique for learning from cases and generating the missing detailed ethical rules needed for reasoning about future similar cases. The newly learned rules are to be added to the agent knowledge base. 
ASP, the purely declarative  non-monotonic reasoning paradigm, was chosen because ethical rules are said to be default rules, which means that they tolerate exceptions. This in fact nominates non-monotonic logics which simulate common sense reasoning to be used for formalizing different ethical conceptions. In addition, there are the many advantages of ASP including it is expressiveness, flexibility, extensibility, ease of maintenance, readability of its code. In addition, the existence of solvers to derive consequences of different ethical principles automatically can help in precise comparison of ethical theories, and makes it easy to validate our models in different situations.
ILP was chosen as a machine learning approach because ILP as a logic-based machine learning approach supports two very important and desired aspects of machine ethics implementation into artificial agents viz. explainability and accountability, ILP is known for its explanatory power, clauses of the generated rules can be used to formulate an explanation for the choice of certain decisions over others; moreover, ILP also seems better suited than statistical methods to domains in which training examples are scarce as in the case of ethical domain.

This Paper is organized as follows: In section \ref{pre} we give a short background, then in Section \ref{state}, we briefly review the state of the art. In Section \ref{approach} we present our approach with examples. Then we conclude with future directions in Section \ref{conclude}.  

\section{Preliminaries}
\label{pre}
\subsection{Answer Set Programming}
ASP is a logic programming paradigm under answer set (or  "stable model") semantics \cite{GelLif88}, which applies ideas of autoepistemic logic and default logic. In ASP, search problems are reduced to computing answer sets, and an answer set solver (i.e., a program for generating stable models) is used to find solutions.
An answer set Program is a collection of rules of the form:  $H\leftarrow A_{1} , \ldots , A_m, not A_{m+1}, \ldots, not A_n$ 
were each of $A_i$'s is a literal in the sense of classical logic. Intuitively the above rule means that if $A_1, \ldots , A_m$ are true and if $A_{m+1}, \ldots , A_n$ can be safely assumed to be false then $H$ must be true. The left-hand side and right-hand side of rules are called \emph{head} and \emph{body}, respectively. A rule with empty body ($n = 0$) is called a \emph{fact}. A rule with empty head is a \emph{constraint}, and states that literals of the body cannot be simultaneously true in any answer set. Unlike other semantics, a program may have several answer sets or may have no answer set. So, differently from traditional logic programming, the solutions of a problem are not obtained through substitutions of variables values in answer to a query. Rather, a program $\Pi$ describes a problem, of which its answer sets represent the possible solutions. For more information about ASP and its applications the reader can refer, among many, \cite{DyoubCG18} and the references therein).
\subsection{Inductive Logic Programming}
ILP \cite{muggleton1991inductive} is a branch of artificial intelligence (AI) which investigates the inductive construction of logical theories from examples and background knowledge. In the general settings, we assume a set of Examples \textit{E}, positive $E^+$ and negative $E^-$, and some background knowledge \textit{B}. An ILP algorithm finds the hypothesis \textit{H} such that $B \bigcup H \models E^+$ and $B \bigcup H \not\models E^-$. The possible hypothesis space is often restricted with a language bias that is specified by a series of mode declarations \textit{M}. A mode declaration is either a head declaration \textit{modeh(r, s)} or a body declaration \textit{modeb(r, s)}, where \textit{s} is a ground literal, this scheme serves as a template for literals in the head or body of a hypothesis clause, where \textit{r} is an integer, the recall, which limits how often the scheme can be used. A scheme can contain special
\textit{placemarker} terms of the form \textit{$\sharp$ type}, \textit{+type} and \textit{-type}, which stand, respectively, for ground terms, input terms and output terms of a predicate \textit{type}. Finally, it is important to mention that ILP has found applications in many areas. For more information on ILP and applications, refer, among many to \cite{MuggletonR94} and references therein.

\section{State of The Art}
\label{state}
Moral decision-making and judgment is a complicated process involving many aspects: it is considered as a mixture of reasoning and emotions. In addition moral decision making is highly flexible, contextual and culturally diverse. Until now it is agreed upon that there is a lack of general theory to guide ethical decision making. Below we briefly review the research work done to model ethical machines using ASP and others using ILP.

LP particularly ASP were used to formalize different ethical conceptions, logical representations help to make ideas clear and highlight differences between different ethical systems.
 
In \cite{ganascia2007modelling}, the authors formalized three ethical conceptions (the Aristotelian rules, Kantian categorical imperative, and Constant's objection) using nonmonotonic logic, particularly Answer Set Programming. In \cite{BerrebyBG17}, authors modeled many ethical theories of the right, and implemented them in ASP. However, their framework assesses the permissibility of an action or a set of actions using different theories of Good and Right separately, i.e. it only permits to judge an action with respect to a single ethical principle. It doesn't handle the conflicting decisions given by different theories, i.e. it doesn't provide a final decision for the agent about what it should do as a result. Pereira and Saptawijaya have proposed the use of different logic-based features for representing diverse issues of moral facets \cite{PereiraS16}. In their formalization, the relationship between the action and its consequences is stated by the programmer rather than inferred, i.e. they are not dynamically linked. Thus, it fails to account for causality and ethical responsibility. Furthermore, because they automatically specify the ethical character of the situation outcome, one needs to write different programs for each case. This is redundant and can lead to inconsistencies. In \cite{CointeBB16}, authors introduced a model that can be used by the agent in order to judge the ethical dimensions of its own behavior and the behavior of others. Their model was implemented in ASP. However, the model is still based on a qualitative approach. Whereas it can define several moral valuations, there is neither a degree of desires, nor a degree of capability, nor a degree of rightfulness. Moreover, ethical principles need to be more precisely defined to capture various sets of theories suggested by philosophers.
Sergot in \cite{sergot2016engineering}, provides an alternative representation to the argumentative representation of a moral dilemma case concerning a group of diabetic persons, presented in \cite{AtkinsonB06}, where the authors used value-based argumentation to solve this dilemma. According to Sergot, the argumentation framework representation doesn't work well and doesn't scale. Sergot proposal for handling this kind of dilemmas is based on Defeasible Conditional Imperatives. The proposed solution was implemented in ASP.

Ethics is more complicated than following a single ethical principle. According to Ross (\cite{ross2002right}), ethical decision making involves considering several Prima Facie duties, and any single-principled ethical theory like Act Utilitarianism is sentenced to fail.
ILP was used by researchers to model ethical decision making in MedEthEx \cite{AndersonAA05}, and EthEl \cite{AndersonA08}. These two systems are based on a more specific theory of prima facie duties viz., the principle of Biomedical ethics of Beauchamp and Childress \cite{beauchamp1991principles}.
In these systems, the strength of each duty is measured by assigning it a weight, capturing
the view that a duty may take precedence over another. Then computes, for each
possible action, the weighted sum of duty satisfaction, and the right action is the one with the greatest sum. The three systems use ILP to learn the relation \textit{supersedes(A1,A2)} which says that action \textit{A1} is preferred over action \textit{A2} in an ethical dilemma involving these choices. 
MedEthEx is designed to give advice for dilemmas in biomedical fields, while EthEl is applied to the domain of eldercare with the main purpose to remind a patient to take her medication, taking ethical duties into consideration. 
GenEth \cite{AndersonA14} is another System that makes use of ILP. GenEth has been used to codify principles in a number of domains relevant to the behavior of autonomous systems.

\begin{table}
	\begin{center}
		
		\footnotesize
		\begin{tabular}{l  l}
			
			\hline
			\textbf{window} w1 & \textbf{} \\
			\hline
			\textbf{Facts} & \textbf{Conclusion}  \\
			\textit{ask(customer,infoabout(productx)).} & \textit{unethical(healthy-way-to-loose-wieght).}\\
			\textit{answer(healthy-way-to-loose-wieght).}& \\
			\textit{not\_SupportEvidence(healthy-way-to-loose-wieght).}& \\
			
			\textbf{Kernal Set} & \textbf{Variabilized Kernal Set}  \\
			unethical(healthy-way-to-loose-wieght) $\leftarrow$  & K1= unethical(V) $\leftarrow$ \\ 
			\qquad answer(healthy-way-to-loose-wieght), & \qquad answer(V),\\
			\qquad not\_SupportEvidence(healthy-way-to-loose-wieght).& \qquad not\_SupportEvidence(V).\\
			\textbf{Running Hypothesis} & \textbf{Support Set}  \\
			$H1$= unethical(V) $\leftarrow$ answer(V).& $H1.supp=\{K1\}$\\
			
			\hline
			
			\textbf{window} w2 & \textbf{} \\
			\hline
			\textbf{Facts} & \textbf{Conclusion}  \\
			ask(customer,infoabout(productY)). & not\_unethical(xxx).\\
			answer(xxx). supportEvidence(xxx).& \\
			
			\textbf{Revised Hypothesis} & \textbf{Support Set}  \\
			$H2$= unethical(X1) $\leftarrow$answer(X1), not\_SupportEvidence(X1)& $H2.supp=\{K1\}$\\
			
			\hline
			
			\textbf{window} w3 & \textbf{} \\
			\hline
			\textbf{Facts} & \textbf{Conclusion}  \\
			ask(customer,infoabout(productZ)). & unethical(WithoutOurProductYouBecomeFat).\\
			answer(WithoutOurProductYouBecomeFat)& \\
			exploitEmotions(WithoutOurProductYouBecomeFat)& \\
			spreadFalseBelief(WithoutOurProductYouBecomeFat)& \\
			
			\textbf{Kernal Set} & \textbf{Variabilized Kernal Set}  \\
			unethical(WithoutOurProductYouBecomeFat) $\leftarrow$  & K2= unethical(X1) $\leftarrow$ \\ 
			\qquad answer(WithoutOurProductYouBecomeFat), & \qquad answer(X1),\\
			\qquad 	exploitEmotions(WithoutOurProductYouBecomeFat), & \qquad exploitEmotions(X1),\\ \qquad spreadFalseBelief(WithoutOurProductYouBecomeFat).& \qquad spreadFalseBelief(X1).\\
			\textbf{Running Hypothesis:}Remains unchanged & \textbf{Support Set:} $H2.supp=\{K1,K2\}$ \\

			\hline
			
			\textbf{window} w4 & \textbf{} \\
			\hline
			\textbf{Facts} & \textbf{Conclusion}  \\
			ask(customer,infoabout(productW)).answer(www). & not\_unethical(www).\\
			
			exploitEmotions(www), not\_spreadFalseBelief(www)& \\
			
			\textbf{Revised Hypothesis} & \textbf{Support Set}  \\
			$H31$= unethical(X1) $\leftarrow$ answer(X1), not\_SupportEvidence(X1)).& $H3.supp=\{K1,K2\}$\\
			$H32$= unethical(X1) $\leftarrow$ answer(X1), spreadFalseBelief(X1). & \\
			\hline
			\textbf{window} w5 & \textbf{} \\
			\hline
			\textbf{Facts} & \textbf{Conclusion}  \\
			ask(customer,infoabout(productR)). & not\_unethical(rrr).\\
			answer(rrr).& \\
			not\_exploitEmotions(rrr), not\_spreadFalseBelief(rrr).& \\
			
			\textbf{Running Hypothesis:}Remains unchanged & \textbf{Support Set:} $H3.supp=\{K1,K2\}$ \\

			\hline
			\textbf{window} w6 & \textbf{} \\
			\hline
			\textbf{Facts} & \textbf{Conclusion}  \\
			ask(customer,infoabout(productS)).answer(sss). & unethical(sss).\\
			
			exploitEmotions(sss). spreadFalseBelief(sss)& \\
			
			\textbf{Revised Hypothesis} & \textbf{Support Set}  \\
			$H31$= unethical(X1) $\leftarrow$ answer(X1), not\_SupportEvidence(X1)).& $H3.supp=\{K1,K2\}$\\
			
			$H32$= unethical(X1) $\leftarrow$answer(X1), spreadFalseBelief(X1),& \\ 
			\qquad \qquad	\qquad \qquad \qquad exploitEmotions(X1). &\\
			\hline

		\end{tabular}
		
		\caption{Example:Input examples and output theory}
	\end{center}
\end{table}

\section{Building the Ethical Agent: Our Approach}
\label{approach}
In this section we present our approach, the application we are considering is an online customer service chatbot. In this work we are concerned only with the ethical reasoning capabilities of our agent, other details related to the complete design of a chatbot are not handled here. The behavior of an ethical online customer service chatbot should be dictated by the codes of ethics and conduct of its company.
Codes of ethics in domains such as customer service are abstract general principles, they apply to a wide range of situations. They are subject to interpretations and may have different meanings in different contexts. There are no intermediate rules that elaborate these abstract principles or explain  how they apply in concrete situations. 
We propose an approach to generate these intermediate rules from interactions with clients through a simplified dialogue. The newly generated rules are to be added to our agent knowledge base, to be used for ethical reasoning of future cases.
Initially our agent will have a very small ethical background knowledge limited to few ethical rules represented by ASP like:
$rule1 = \{unethical(V) \leftarrow not\_correct(V), answer(V).\}$
which says that it is unethical to give incorrect information to the customers. The missing ethical rules are learned by our agent incrementally overtime through interactions with clients.
During the training phase, the trainer enters a series of sentences in the form of requests and responses through the keyboard  simulating a customer service chat point conversation, along with the ethical evaluation of the responses in each scenario. The first step is to convert the natural language sentences to the syntax of ASP (e.g. refer \cite{PendharkarG19}). The system remembers the facts about the narratives given by the trainer and learns to form ethical evaluation rules according to the facts given in the story context (\textit{C}) and background knowledge (\textit{B}). For learning the ethical rules (\textit{H}) needed for dictating the ethical behavior of our agent, we use the state of the art ILP tool ILED \cite{KatzourisAP15}.
In the test phase, the agent uses both \textit{B \& H} to respond to the client request avoiding unethical practices. The goal is to recognize unethical responses from combinations of case' facts.

To illustrate our approach, let us consider the following scenarios were we want to teach our customer service chatbot that it is natural to highlight or exaggerate the best features of a product or a service, but this practice crosses the ethical line when it comes to messaging that misleads the customer, such as marketing a product as a healthy way to lose weight when there isn’t significant evidence to support such a claim. Or that appealing to emotions is an effective way to reach customers, but it is unethical to intentionally evoke emotions like rage, fear, sadness, etc. to manipulate customers. And it is unethical to spread a false belief that a certain product or service can only save them. These practices violate honesty and truthfulness. To do so the trainer will provide the system with different positive and negative examples, table 1 demonstrate the learning process. 

The system will start constructing hypothesis from the first available case (c1). The generated hypothesis (rule) will be added to the agent knowledge base. When a new case (c2) arrive, the system will check whether the new case is covered by the running hypothesis. If not, it will start the revision process to update the running hypothesis (rule) to a new rule that cover the new case (see table1). Considering the following mode declarations serving as patterns for restricting the hypotheses search space: $M=<M_h,M_b>$ with head declarations $M_h= \{unethical(V)\}$ and body declarations $M_b=\{not\_supportEvedence(V), \allowbreak spreadFalseBelief(V), exploitEmotions(V)\}$, where V denotes that the arguments of the predicates are variables.
In our preliminary experiment we used similar setting to those used in (\cite{KatzourisAP15}). We created a small set of examples which we stored in a 'mongodb' database, then we run ILED to learn incrementally from this dataset.

\section{Conclusions And Future Directions}
\label{conclude}
Combining ASP with ILP for modeling ethical agents provides many advantages: increases the reasoning capability of our agent; promotes the adoption of hybrid strategy for modeling ethical agents; allows the generation of rules with valuable expressive and explanatory power which equips our agent with the capacity to explain the reasons behind its actions. In other words, our method supports transparency and accountability of such models, which facilitates instilling confidence and trust in our agent. 
Furthermore, in our opinion and for the sake of transparency, machines ethical behavior should be guided by explicit ethical rules determined by competent judges or ethicists or through consensus of ethicists. Our approach provides support for developing these ethical rules. 

ILP algorithms, unlike neural networks, output rules which are easily understood by people. Lack of intuitive descriptions makes it hard for users to understand and verify the underlying rules that govern the model. Also, statistical methods cannot produce
a justification for a prediction they compute. Furthermore, if background knowledge is extended, then the entire model needs to be re-learned. ILP particularly appropriate for tasks in which the comprehensibility of the generated knowledge is essential. Moreover, in an ill-defined domain like the ethics domain, it is infeasible to define abstract codes in precise and complete enough terms to be able to use deductive problem solvers to apply them correctly. A combination of deductive (rule-based) and inductive (case-based learning) is needed.

As a future work, we would like to test our agent in a real chat scenario. Finally, as another future direction we would like to investigate the possibility of judging the ethical behavior from a series of related chat sessions.



\bibliographystyle{eptcs}

\end{document}